\documentclass{article}

\usepackage{arxiv}

\usepackage[utf8]{inputenc} % allow utf-8 input
\usepackage[T1]{fontenc}    % use 8-bit T1 fonts
\usepackage{hyperref}       % hyperlinks
\usepackage{url}            % simple URL typesetting
\usepackage{booktabs}       % professional-quality tables
\usepackage{amsfonts}       % blackboard math symbols
\usepackage{nicefrac}       % compact symbols for 1/2, etc.
\usepackage{microtype}      % microtypography
\usepackage{graphicx}
\usepackage{amsmath}
\usepackage{float}
\usepackage[numbers]{natbib}
\graphicspath{ {./images/} }

\title{Synthics: synthetic physics-like datasets for machine learning}

\author{
 Jari Vepsäläinen \\
  Mechatronics\\
  Department of Energy and Mechanical Engineering\\
  School of Engineering\\
  Aalto University\\
  Finland\\
  \texttt{jari.vepsalainen@aalto.fi} \\
}

\begin{document}
\maketitle
\begin{abstract}
Representative data is fundamental in machine learning, as limited data hinders generalisation.
Collecting sufficient real-world samples is often infeasible.
Synthetic data generation offers a practical solution, but only if the generated data faithfully reflects the structure of real observations.
In this paper, a method for generating synthetic regression datasets that structurally resemble physics equations from a given equation corpus is presented.
The approach uses a Bayesian Probabilistic Context-Free Grammar to capture the underlying algebraic structure of the corpus, from which novel equations are sampled.
To ensure the generated inputs lie within a physically meaningful domain, the applicability domain is characterised for each equation through non-intrusive probing, also recovering inter-variable constraints.
Input sampling further mimics realistic experimental conditions by drawing from random sub-ranges of the valid domain with mixed uniform and truncated normal distributions.
The generated data is statistically validated against the Feynman equation corpus using Kolmogorov–Smirnov tests.
The generated equations match the corpus on all of the eight studied structural features, compared to only two for an unsmoothed purely probabilistic grammar, demonstrating that the Bayesian prior is essential for structural fidelity given the size of the corpus.
In a downstream hyperparameter-tuning task, a gradient-boosted regressor tuned on the synthetic data picks, on average, the 6th-best configuration out of 20 on real data, matching the result of tuning on real data itself and substantially outperforming random expression trees (10th) and noise (19th).

\end{abstract}


\section{Introduction}

Machine learning models are only as good as the data they are trained on, yet many engineering applications are data-scarce.
Unlike large language models, which draw on the vast text available online, datasets of comparable scale for physical systems are rarely available, and acquiring representative data is often unrealistically expensive and time-consuming.
Each new sample typically requires running an experiment that is slow, costly, and often confined to narrow operating regimes by safety or regulatory constraints.

Synthetic data generation offers a possibility to overcome data scarcity, but its usefulness depends on how faithfully the generated data reflects the structure of real observations.
Unlike data augmentation, which perturbs existing samples to add diversity, synthetic data creation produces entirely novel samples that emulate the target domain and allows a more broad exploration of the feature space.
The challenge is not the quantity of synthetic data but its quality.
With sufficiently realistic synthetic data, models can learn to generalise to real-world tasks without ever being trained on real data.
For example,TabPFN is a tabular foundation model meta-learned across millions of synthetic datasets, and it outperforms tuned gradient-boosted trees~\cite{hollmann2025accurate}.

This paper introduces a method for generating synthetic datasets that preserve the structural and mathematical properties of given physics equations.
The method learns a Bayesian Probabilistic Context-Free Grammar (B-PCFG) from a corpus of physics equations and samples novel equations from the learned distribution.
These equations are turned into datasets through constrained input sampling and applicability-domain estimation, so that the generated points respect each equation's mathematical constraints, for example requiring $x_1 - x_2 \geq 0$ wherever $\sqrt{x_1 - x_2}$ appears.
The generated equations are then validated against the corpus with Kolmogorov--Smirnov tests.
The practical utility of the generated datasets is demonstrated in a hyperparameter tuning task where it is shown that tuning on the synthetic data guides hyperparameter selection for a machine learning task better than tuning on random expression trees or noise.

The contributions of this paper are as follows:
\begin{enumerate}
    \item A constrained input sampling framework for physics equations.
    The proposed method combines uniform and truncated-normal sampling while performing expression-specific domain estimation to avoid invalid evaluations caused by operators such as square roots and logarithms.
    
    \item A grammar-based framework for generating synthetic regression datasets from physics equations.
    Instead of sampling equations from uniformly random expression trees, the proposed Synthics framework learns a B-PCFG from a corpus of physics equations and generates novel expressions from the learned distribution.
    This preserves structural characteristics such as operator frequencies, expression depth, and variable interaction patterns while maintaining syntactic novelty.
    
    \item Statistical and practical validation of methodology.
    The similarity between generated and reference equation distributions is evaluated using Kolmogorov--Smirnov-based analysis.
    It is also demonstrated that models trained on the synthetic data can guide hyperparameter selection in a machine-learning task. 
\end{enumerate}

\section{Literature review}

\subsection{Synthetic data for machine learning}
Machine learning models trained on rich datasets routinely outperform those trained on small ones, yet in many engineering domains data is scarce.
\citet{alzubaidi2023survey} survey the deep-learning techniques developed to tackle cases with scarce data, such as transfer learning, self-supervised learning, generative models, and physics-informed networks.
They emphasise their role in domains ranging from structural health monitoring to fluid mechanics where data acquisition is costly or limited.

Synthetic data generation is one practical response.
The utility of synthetic data depends less on its volume than on how faithfully the generating distribution reflects real structure.
\citet{goyal2024synthetic} review the generative-AI techniques used to produce it, ranging from generative adversarial networks and variational autoencoders to large language models, and list recurring limitations such as heavy computation, unstable training, and weak privacy guarantees.
Even with these limitations, synthetic data has been shown to improve performance in various applications, such as in healthcare \citep{basri2025useful}, autonomous driving \citep{tremblay2018training}, and robotic manipulation \citep{tobin2017domain}.
\citet{tobin2017domain} trained object detectors entirely on simulated images with randomised textures and lighting and transferred them to real scenes with centimetre-level accuracy, finding that sufficient simulator variability makes the real world appear to the model as merely another variation.
\citet{tremblay2018training} extended this idea to vehicle detection and showed that training on non-photorealistic synthetic data alone yields competitive detectors.

For numerical value prediction, \citet{hollmann2025accurate} introduced TabPFN, a tabular foundation model meta-learned across millions of synthetic datasets that outperforms tuned gradient-boosted trees on small tasks within seconds.
\citet{zhang2025mitra} showed that with a curated mixture of synthetic priors can improve tabular foundation models without changing the underlying architecture.
For engineering specifically, the right structure is dictated by the underlying physics.
\citet{karniadakis2021physics} review the field of physics-informed machine-learning and argue that embedding physical laws into the learning process is essential when data alone is insufficient.
Thus, both the integration of physics-based models with data-driven approaches and the generation of physics-informed synthetic data should be considered complementary strategies to address data scarcity in engineering applications.

\subsection{Meta-learning and model selection}
A particularly impactful application of synthetic data is meta-learning and model selection, where the task is to choose the most suitable algorithm or hyperparameter configuration for a given task.
\citet{rice1976algorithm} originally formalised algorithm selection as a learning problem in which the best algorithm for a given dataset is predicted from problem features.
\citet{smith2008cross} surveyed how the principles presented by Rice have been applied in practice, and \citet{vanschoren2018meta} review the broader meta-learning field, which includes algorithm selection but also extends to few-shot learning, transfer learning, and other paradigms.
Infrastructure for such studies has been provided by \citet{vanschoren2014openml}'s OpenML platform, and systems such as auto-sklearn \citep{feurer2015efficient} apply meta-learning to automate model selection at scale with an ensemble modelling approach.
However, most approaches rely on large datasets and excessive computation due to the fitting of multiple candidate metamodels.
It would be more beneficial to train general classifiers on synthetic data that can predict the best model and configuration for a new task without needing to fit multiple models on it.
\citet{basri2025useful} evaluate synthetic data for hyperparameter tuning and find a strong correlation between prediction accuracy on synthetic and real data.
Thus, there is evidence of the utility of synthetic data for meta-learning and model selection tasks, which motivates the study of synthetic data generators that can produce data with the right structure for such tasks.

\subsection{Grammar-based generation of equation structure}
Grammar-based equation generation is a common approach to produce syntactically valid expressions with interpretable structure.
\citet{kusner2017grammar} introduced the Grammar VAE, which encodes and decodes expressions through the parse trees of a context-free grammar so that every generated expression is syntactically valid, raising the proportion of valid outputs and yielding a more coherent latent space for downstream optimisation, although its grammar is fixed in advance and its rule probabilities are not fitted to a target corpus.
Probabilistic context-free grammars instead attach probabilities to the production rules, and \citet{brence2021probabilistic} used these as a prior over the space of equations to guide equation discovery, expressing the parsimony principle through rule weights and laying the foundations for Bayesian treatments.
This work builds on these foundations and extends them to synthetic data generation rather than equation discovery, retaining an interpretable probabilistic grammar while regularising its production probabilities with a Dirichlet prior so that even a small corpus yields a diverse structural distribution rather than collapsing onto its most frequent rules.
\citet{meznar2023efficient} showed that a learned hierarchical generator can outperform such hand-crafted grammars, demonstrating that rule statistics estimated from data carry genuine structural information, although its neural latent space sacrifices the transparency of an explicit grammar.
Equation discovery is also referred to as symbolic regression \cite{angelis2023artificial, makke2024interpretable}.
\citet{udrescu2020ai} introduced AI Feynman, which combined neural-network fitting with physics-inspired simplifications to recover all 100 equations of the Feynman benchmark.
Recent symbolic regression methods are based on deep learning \citep{biggio2021neural, kamienny2022end, shojaee2023transformer}, but as \citet{sato2025reproduction} show, the effective search space they utilise is narrowed by reproduction bias, with most generated expressions copied from the training distribution rather than freshly synthesised.
Equation structure and discovery have therefore been thoroughly investigated in the literature, but they are only half of the story for synthetic data generation.
The other half is ensuring that the generated equations are sampled with inputs that yield valid outputs.

\subsection{Applicability domains and physically meaningful sampling}
Ensuring valid inputs is a separate problem, because uniform sampling across a nominal range can produce singular, undefined, or physically impossible outputs.
The design-optimisation literature studies this as feasible-region or applicability-domain identification.
\citet{chen2017active} proposed Active Expansion Sampling to learn feasible domains over an unbounded input space and proved a constant probabilistic bound on the misclassification loss within the explored region, independent of the number of iterations.
\citet{metta2020novel} used a neural-network surrogate with jackknife variance estimates to sample adaptively near the feasible-region boundary.
\citet{straus2021constrained} showed that adding inequality constraints capturing interdependencies between variables sharply reduces the sampling domain and improves surrogate accuracy.
\citet{liang2025chance} show that standard generative models often violate hard physical constraints and propose Chance-Constrained Flow Matching, which integrates stochastic optimisation into the sampling process to guarantee feasibility while maintaining high-fidelity generation.
Each of these efforts treats the valid-input problem on its own, without coupling it to the structural generation of the equations being sampled.
The equations generated in this work are closed-form and inexpensive to evaluate, and many of their feasibility conditions follow directly from the expression, such as the non-negativity required beneath a square root.
A simpler, non-intrusive sampling procedure is therefore sufficient, rather than the more sophisticated approaches presented in the literature, which target the harder setting of an unknown feasible region under an expensive black-box model.

\subsection{Research gap}
The literatures surveyed above developed largely independently.
Synthetic-data and meta-learning research has focused on volume and distributional fidelity but rarely on interpretable structure.
Symbolic regression moves from data to equation, the opposite direction of the present task.
Grammar-based methods are powerful but have not been combined with corpus-learnt priors and applicability-domain probing to produce machine-learning-ready datasets.
Applicability-domain estimation has been treated in isolation from the structural generation of the equations being sampled.

No existing approach learns equation structure from a real physics corpus and samples physically valid inputs.
This paper closes that gap by combining a Bayesian probabilistic context-free grammar that learns structural rules from a real physics corpus with non-intrusive applicability-domain probing and constrained input sampling.
The synthetic data is validated structurally against the corpus using Kolmogorov--Smirnov tests and practically through a hyperparameter ranking task.

\section{Methods}

\subsection{Synthics Workflow}
\begin{figure}[H]
  \centering
  \includegraphics[width=0.65\linewidth]{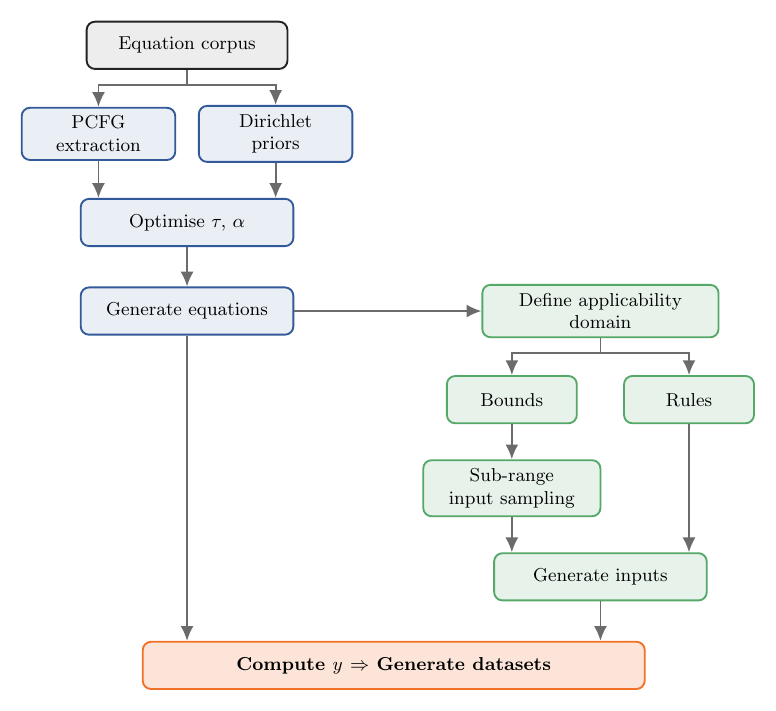}
  \caption{The Synthics pipeline. From an equation corpus, a probabilistic
  context-free grammar is extracted and combined with Dirichlet priors; the
  hyperparameters $(\tau, \alpha)$ are jointly tuned, and the resulting B-PCFG
  generates new equations. Each generated equation defines its own
  applicability domain (bounds and dependency rules), from which inputs are
  drawn by sub-range sampling. Combining equations and inputs yields the final
  synthetic regression datasets.}
  \label{fig:pipeline}
\end{figure}

Figure~\ref{fig:pipeline} shows the Synthics workflow.
Dataset generation is split into two stages that meet at a final composition step: equation generation and input generation.
A Bayesian probabilistic context-free grammar (B-PCFG) is used to extract the structure from the corpus.
The PCFG captures the structural distribution of the corpus by counting how frequently each production rule is used.
The Bayesian extension applies a Dirichlet prior to these counts, which prevents the grammar from collapsing onto its most frequent rules under a small corpus and improves structural diversity.
The B-PCFG matches the structural distribution of the corpus by tuning two hyperparameters $\alpha$ and $\tau$.

The second stage operates per generated equation: a non-intrusive probing step characterises the applicability domain, yielding a per-variable bounding box together with a set of inter-variable dependency rules.
Inputs are then drawn from a random sub-range of that domain under a mixed uniform and truncated-normal distribution.
Uniform and truncated-normal distributions are selected as the two most common parametric forms, with the normal truncated to the valid domain to avoid invalid samples.
The extracted rules act as a rejection filter so that every sample satisfies the equation's mathematical constraints.
At the final step, each generated equation is evaluated on its sampled inputs to produce one regression dataset, and iterating over many equations yields the full synthetic dataset corpus.

\subsection{Equation Corpus}
The data utilised in this study comprises the 100 foundational physics equations compiled from the renowned Feynman Lectures on Physics \cite{feynman1965feynman}.
This corpus was established as a benchmark by Udrescu and Tegmark \cite{udrescu2020ai} for their 'AI Feynman' project, which introduced a novel symbolic regression framework designed to discover mathematical formulas from raw data.
The complete dataset is publicly accessible via the Feynman Symbolic Regression Database \cite{FeynmanDatabase2020}.

\subsection{Bayesian Probabilistic Context-Free Grammar (B-PCFG)}
To generate symbolic equations that resemble real physical laws, the equation structure is modelled using a Probabilistic Context-Free Grammar (PCFG).
In this framework, equations are represented as expression trees, where each node corresponds to a mathematical operation such as addition, multiplication, or a trigonometric function.
The grammar defines how one operator can expand into combinations of other operators, variables, or constants.
For example, a multiplication node may expand into two subexpressions, while a sine node expands into a single argument.
Rather than assigning these expansions uniformly at random, the PCFG learns probabilities from a corpus of real equations.

Each production rule therefore has an associated probability:
\begin{equation}
    A \rightarrow \beta
\end{equation}
where $A$ denotes an operator type and $\beta$ denotes its expansion pattern.

The probability of generating an equation tree is obtained by multiplying the probabilities of all applied rules.
As a result, equation structures that resemble those observed in real physical laws become more likely to be generated.

The grammar is learned from the Feynman equation corpus.
Each equation is parsed into a symbolic expression tree using \textsc{SymPy}, where internal nodes represent operators and leaves represent either variables or constants.
For every operator type, the frequency of each expansion pattern is counted across the corpus.
These counts are then converted into probabilities, allowing the grammar to capture common structural characteristics of physical equations, including operator usage, nesting depth, and variable interactions.

\begin{figure}[H]
  \centering
  \includegraphics[width=0.9\linewidth]{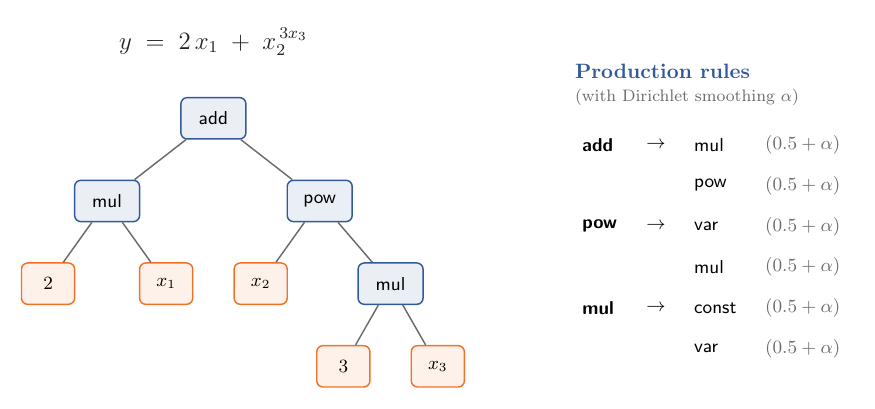}
  \caption{An example equation expression tree and the corresponding Bayesian-smoothed production rules.
  Each rule's probability is the maximum-likelihood frequency plus a Dirichlet pseudo-count $\alpha$,
  which prevents the grammar from collapsing onto its most frequent rules under a small corpus.
  The $\alpha$ is actually incorporated into equation~\ref{eq: alpha} but for visual simplicity it is shown here as an additive term.
  }
  \label{fig:pcfg}
\end{figure}

A direct frequency estimate can become overly dominated by the most common equation patterns, especially when the corpus is relatively small.
This may suppress rare but structurally meaningful expressions.
To address this, Bayesian smoothing is applied using a Dirichlet prior.
Conceptually, this adds a small pseudo-count to every possible rule before estimating probabilities.
The resulting rule probability becomes:

\begin{equation}
    \hat{\theta}_{A,\beta}
    =
    \frac{c_{A,\beta} + \alpha}
    {C_A + \alpha K_A},
    \label{eq: alpha}
\end{equation}

where $c_{A,\beta}$ is the observed count of a rule, $C_A$ is the total number of rules for operator $A$, $K_A$ is the number of possible expansions, and $\alpha$ controls the smoothing strength.
Small values of $\alpha$ closely follow the original corpus frequencies, while larger values produce a more uniform distribution over rules.
In practice, this smoothing improves structural diversity and prevents dominant patterns from overwhelming the generation process.
Figure~\ref{fig:pcfg} shows an example parse tree along with the corresponding production rules and their smoothed probabilities, illustrating how the grammar captures the structure of a sample equation.

New equations are generated recursively from the learned grammar by repeatedly sampling production rules according to their Bayesian-smoothed probabilities.
To prevent excessively deep expression trees, the probability of terminating a branch increases gradually with recursion depth:
\begin{equation}
    p_{\text{leaf}}(d)
    =
    1 - \exp\left(-\frac{d}{\tau}\right),
\end{equation}
where $\tau$ controls the typical tree depth.

Leaf nodes are sampled as either variables or constants, and variables are assigned symbolic names such as $x_1$, $x_2$, and so forth.
Equations that contain fewer than two free variables are discarded and regenerated.
The hyperparameters $\alpha$ and $\tau$ are selected by comparing statistical properties of the generated equations against the original corpus using Kolmogorov--Smirnov tests.
These properties include expression depth, operator counts, and variable usage distributions.

\subsection{B-PCFG Hyperparameter Optimisation}
The B-PCFG generator is controlled by two hyperparameters: the Dirichlet concentration parameter $\alpha$, which governs the strength of the Bayesian smoothing applied to the production-rule probabilities, and the soft-forcing temperature $\tau$, which controls the expected depth of the generated expression trees.
The two interact non-trivially, since $\alpha$ reshapes the distribution over rules while $\tau$ independently regulates the rate of soft termination, so they must be tuned jointly rather than in isolation.

Although both hyperparameters take integer values over bounded ranges, an exhaustive grid search is impractical, because every candidate requires generating $N_{\text{gen}} = 500$ equations and computing eight Kolmogorov--Smirnov (KS) tests against the corpus.
The choice of $(\alpha, \tau)$ is therefore treated as a sample-efficient sequential optimisation problem rather than an exhaustive sweep, searching over the integer space $\alpha \in \{1, \ldots, 50\}$ and $\tau \in \{1, \ldots, 25\}$.

Each candidate is scored by a composite fitness that rewards distributional agreement with the corpus across the eight structural features.
Writing $p_i$ and $D_i$ for the KS $p$-value and statistic of feature $i$, and $n_{\text{pass}}$ for the number of features passing at $p > 0.05$, the fitness is
\begin{equation}
    f(\alpha, \tau) = 5 \sum_{i=1}^{8} p_i \;-\; \sum_{i=1}^{8} D_i \;+\; 15\, n_{\text{pass}},
    \label{eq:fitness}
\end{equation}
so that both graded similarity and the number of outright passes are rewarded.
These weights were selected by iterative experimentation, which aimed to find weights that yield robust optimisation convergence.
The fitness $f$ is maximised with Optuna~\cite{akiba2019optuna} using its Tree-structured Parzen Estimator (TPE) sampler~\cite{bergstra2011algorithms}.
\citet{bergstra2011algorithms} proposed TPE and showed that such sequential model-based methods outperform random and manual search on hard tuning problems.
The search runs for $100$ trials under a fixed random seed for reproducibility.
The best configuration $(\alpha^*, \tau^*)$ is then re-evaluated on a larger sample of $1000$ generated equations, from which the per-feature KS pass/fail results reported in the Results are obtained.

\subsection{Applicability Domain}
A key challenge in generating synthetic regression datasets from algebraic equations is ensuring that the sampled inputs produce well-defined, finite outputs.
Many physically meaningful equations contain singularities or restricted domains.
For instance, the relativistic Doppler equation $\omega = \omega_0 / (1 - v/c)$ requires $v < c$ to avoid a singularity in the denominator and $c\ne0$, the diffraction
condition $\theta = \arcsin(\lambda / nd)$ requires $\lambda < nd$ for the arcsine to be defined, and the Coulomb force law $F = q_1 q_2 / (4\pi\varepsilon r^2)$ diverges as $r \to 0$ when the
distance between charges approaches zero.
Naively sampling inputs uniformly from $[0, 1]^n$ would produce a large proportion of invalid outputs which pollute the data and cause numerical issues downstream.

\begin{figure}[H]
  \centering
  \includegraphics[width=0.4\linewidth]{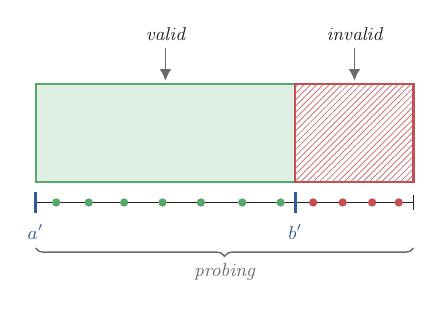}
  \caption{Applicability-domain specification for one variable. The probing
  range is sampled, each point evaluated, and labelled valid or invalid. The
  domain bounds $a', b'$ are the empirical range of the valid points.}
  \label{fig:ad}
\end{figure}

Therefore, it is important to characterise the applicability domain of each equation before sampling.
Here an empirical numerical probing is used for the characterisation. 
A set of $N_{\text{probe}} = 1000$ random points is drawn uniformly from $[0,1]^n$ and evaluated, as shown in Figure~\ref{fig:ad}.
A point is labelled valid if its output is finite and within a physically meaningful magnitude range, specifically $|y| \in (10^{-12}, 10^8)$.
From the valid points, a per-variable bounding box is extracted by taking the 5th--95th percentile of valid input values for each variable.
These validity observations are used to generate pairwise and trivariate dependency rules of the
form:

\begin{align}
    x_i &< \theta_{ij} \cdot x_j \label{eq:ratio_rule}\\
    x_i &< \theta_{ijk} \cdot x_j \cdot x_k \label{eq:product_rule}\\
    x_i &< \theta_{ij} \cdot x_j^2 \label{eq:power_rule}
\end{align}

where the threshold $\theta$ is estimated as the 95th percentile of the corresponding ratio among valid probe points.
Each candidate rule is evaluated by its balanced accuracy in separating valid from invalid inputs, and only rules exceeding a minimum accuracy of 0.75 are retained.
This procedure recovers physically interpretable constraints automatically and without any domain knowledge. For example, the relativistic
constraint $v < c$ is recovered as a ratio rule $x_i < 0.94 \cdot x_j$, and the diffraction constraint $\lambda < nd$ is recovered as a product
rule $x_i < 0.93 \cdot x_j \cdot x_k$.

\subsection{Input Sampling}
\begin{figure}[H]
  \centering
  \includegraphics[width=\linewidth]{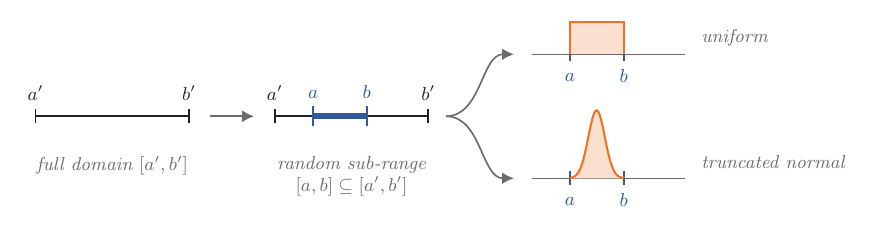}
  \caption{Input sampling within a variable's applicability domain.
  A sub-range $[a, b]\subseteq[a', b']$ is drawn at random, and samples within
  it are then generated under either a uniform or a truncated-normal distribution.}
  \label{fig:input-sampling}
\end{figure}

Given the characterised domain, inputs for each synthetic dataset are sampled using a rejection-sampling scheme that combines distribution diversity with domain constraints.
For each variable, a random sub-interval $[a_j, b_j] \subset [\ell_j, u_j]$ is drawn uniformly within the bounding box $[\ell_j, u_j]$.
The variable is then assigned either a uniform distribution $\mathcal{U}(a_j, b_j)$ or a truncated normal distribution $\mathcal{N}(\mu_j, \sigma_j)$ clipped to $(a_j, b_j)$.
The ratio of uniform to normal variables is controlled by a hyperparameter $r_{\text{uniform}} \in [0, 1]$.
Figure~\ref{fig:input-sampling} illustrates the sampling process, showing how random sub-ranges are drawn within the applicability domain and how samples are generated under either distribution.

This design reflects the observation that real experimental data rarely covers the full theoretically valid input range.
Instead, systems operate within calibrated sub-ranges, and physical processes often
concentrate near a characteristic operating point.
The mixed uniform and truncated normal sampling with random sub-ranges mimics this diversity across datasets while respecting the physical constraints encoded in the dependency rules.

Candidate points are drawn in batches of $5 \times N_{\text{samples}}$ and filtered by the dependency rules as a post-hoc rejection step, repeating until $N_{\text{samples}}$ valid points are collected.
If fewer than 20 probe points were valid during domain characterisation, which would indicate a pathological equation, the equation is discarded entirely and not used for dataset generation.

\subsection{Structural Validation: Kolmogorov–Smirnov Tests}
To assess whether the B-PCFG generator produces equations that are structurally representative of the Feynman corpus, the distribution of eight structural features is compared between the corpus equations and a set of $N_{\text{gen}} = 500$ generated equations using two-sample Kolmogorov--Smirnov (KS) tests.
Each equation is parsed into an expression tree by \textsc{SymPy} and the following scalar features are extracted:

\begin{itemize}
    \item \textbf{Depth} — the length of the longest path from the root to any leaf node.
    \item \textbf{Number of operators} — total count of internal operator nodes in the tree.
    \item \textbf{Number of unique variables} — the number of distinct free symbols appearing in the expression.
    \item \textbf{Number of constants} — the number of numeric constant leaves.
    \item \textbf{Unary ratio} — the fraction of operators that take a single argument (e.g.\ $\sin$, $\exp$, $\sqrt{\cdot}$).
    \item \textbf{Average operator depth} — the mean depth of all operator nodes.
    \item \textbf{Average leaf depth} — the mean depth of all leaf nodes.
    \item \textbf{Branching factor} — the mean number of children per operator node.
\end{itemize}

The Kolmogorov–Smirnov (KS) test is a non-parametric statistical method used to compare distributions by the maximal distance between two cumulative distribution functions.
For each feature $f$, let $\{c_1, \ldots, c_N\}$ and $\{g_1, \ldots, g_M\}$ denote the feature values for the corpus and
generated equations respectively.
The two-sample KS statistic is:

\begin{equation}
    D_{N,M} = \sup_{x} \left| F_N(x) - G_M(x) \right|,
\end{equation}

where $F_N$ and $G_M$ are the empirical cumulative distribution functions of the corpus and generated samples.
A large $D_{N,M}$ indicates that the two distributions differ substantially.
The associated $p$-value tests the null hypothesis that both samples are drawn from the same underlying distribution.
A feature passes the test when its $p$-value is at least the chosen threshold of $0.05$, indicating that the generated equations are statistically indistinguishable from the corpus on that feature.
The KS test is chosen because it is non-parametric and sensitive to differences in shape, location, and spread without assuming any particular distributional form.
This was deemed appropriate given that structural features such as tree depth follow discrete, heavily right-skewed distributions.

\subsection{Practical Validation: Hyperparameter Ranking Task}

The purpose of synthetic datasets is to assist in meta-learning and machine learning tasks, to improve generalisation and performance. 
Here, this is validated by a hyperparameter tuning experiment, where Gradient Boosting Regressors (GBRs) are trained on synthetic datasets generated by the optimised B-PCFG, random expression trees, and noise.
The goal is to see whether tuning with Synthics data outperforms tuning with random trees and noise, and how close it gets to tuning with the real corpus data.

A hyperparameter space of $4 \times 5 = 20$ configurations is studied, defined by:
\begin{equation*}
    \text{max\_depth} \in \{2, 3, 4, 5, 6\} \times \text{learning\_rate} \in \{0.01, 0.05, 0.1, 0.3\}.
\end{equation*}

The comparison is made across four dataset types, which will be referred to as \textit{sources}.
Each is turned into a dataset with the previously presented input sampling and applicability domain.
The 100 Feynman equation corpus is split 80/20, such that:
\begin{itemize}
    \item \textbf{Corpus:} Use the 20 test equations for training and testing, to establish a real-data baseline.
    \item \textbf{Synthics (B-PCFG):} Learn grammar from the training 80 equations. 
    Parameters $\alpha$ and $\tau$ are tuned on the training equations.
    \item \textbf{Random trees:} Sample equations and inputs from randomly generated expression trees.
    \item \textbf{Noise:} Sample inputs the same way as for the others, but set the target to pure noise independent of $X$.
\end{itemize}
Thus, Synthics, Random trees and Noise generators never see the test equations.
They try to learn the best ranking of the hyperparameter configurations based on their own training data.

Evaluation is carried out for each source by training a GBR on each of the 20 hyperparameter configurations and scoring on the test set using $R^2$.
Synthics, Random trees and Noise generators are used to produce 80 novel datasets that are referred to as tasks, while for the Corpus source the 20 test equations themselves are the tasks.
For each task, $n = 300$ points are sampled, split 70/30 into train/test sets.

For each source, per-task $R^2$ is aggregated for each configuration across all of that source's tasks.
Based on the average rank across tasks, the 20 configurations are then ordered into a ranking, denoted $\text{ranking}_S$.
The aggregation is done by mean within-task rank (Borda count) rather than mean $R^2$ because large negative $R^2$ outliers can pollute the average.
Borda count was originally invented for voting systems to aggregate ranked preferences \cite{emerson2013borda}.

The rankings are then compared with the real-data ranking $\text{ranking}_{\text{real}}$ using Spearman's rank correlation coefficient $\rho_S$.
    \begin{equation*}
        \rho_S = \text{Spearman}(\text{ranking}_S, \text{ranking}_{\text{real}})
    \end{equation*}
This serves as the main metric of interest because it assesses the efficacy of synthetic data in guiding hyperparameter tuning for real-world applications.
However, $\rho_S$ is a relative metric that only captures the ordering of configurations, not the magnitude of performance differences.
Therefore, a second metric is also reported, which is the average $R^2$ loss compared to the real-optimal configuration when applying the synthetic-optimal configuration to the real tasks.
This is referred to as the \textit{regret} of a given source.
Regret is measured by comparing the $R^2$ of the best configuration determined by $\max(\text{ranking}_S)$ to the best configuration determined by the test equations, which is $\max(\text{ranking}_{\text{real}})$.

To account for the stochastic nature of the experimental pipeline, all experiments are repeated across five distinct random seeds per iteration.
For each seed, the data is subjected to a new 80/20 train/test split, synthetic data generation, and point resampling.
This iterative process yields a distribution of $\rho_S$ and regret values for each data source to quantify the variability and robustness of the results.
The final results are reported as the mean and standard deviation of $\rho_S$.

\section{Results}

\subsection{Structural Fidelity of B-PCFG}
The Optuna search converged after $100$ trials to $\alpha^* = 44$ and $\tau^* = 6$.
The outcome depends on the random seed, since both grammar extraction and equation generation are stochastic, but across different seeds the optimiser consistently recovers $(\alpha, \tau)$ values that pass most or all of the structural KS tests.

The generator is evaluated by comparing the structural properties of $1000$ generated equations against the $100$-equation Feynman corpus.
Each of the eight structural features is compared with a two-sample Kolmogorov--Smirnov (KS) test, and a feature passes the test when $p > 0.05$, meaning the null hypothesis of a shared distribution between the corpus and generated samples cannot be rejected.
To isolate the effect of the Bayesian smoothing, the standard PCFG is generated with the same temperature $\tau = 6$, so that the concentration parameter $\alpha$ is the only difference between the two generators.

With smoothing applied, the B-PCFG passes all eight KS tests, whereas the standard PCFG passes only two (Table~\ref{tab:ks}).
The distribution overlays in Figures~\ref{fig:pcfg_validation} and~\ref{fig:bpcfg_validation} found in the appendix show that the standard PCFG concentrates probability on shallow, structurally simple trees and underrepresents the depth, operator count and constant count of the corpus, while the B-PCFG recovers the corpus-level spread across every feature.
The agreement is nonetheless far from exact.
Several features pass only marginally, with the number of constants clearing the threshold at just $p = 0.086$, and the corpus and generated histograms remain visibly distinct in places.

\begin{table}[H]
\centering
\caption{Per-feature two-sample Kolmogorov--Smirnov results comparing $1000$ generated equations with the $100$-equation Feynman corpus, for the standard PCFG and the optimised B-PCFG ($\alpha^* = 44$). Both generators use $\tau = 6$, so the only difference is the Bayesian smoothing. A feature passes when $p > 0.05$.}
\label{tab:ks}
\begin{tabular}{lcccc}
\toprule
 & \multicolumn{2}{c}{PCFG} & \multicolumn{2}{c}{B-PCFG} \\
\cmidrule(lr){2-3} \cmidrule(lr){4-5}
Feature & $D$ & $p$ & $D$ & $p$ \\
\midrule
Depth                & 0.299 & 0.000 & 0.077 & 0.631 \\
Operators            & 0.381 & 0.000 & 0.126 & 0.104 \\
Unique variables     & 0.228 & 0.000 & 0.126 & 0.104 \\
Constants            & 0.321 & 0.000 & 0.130 & 0.086 \\
Unary ratio          & 0.115 & 0.169 & 0.104 & 0.264 \\
Avg.\ operator depth & 0.350 & 0.000 & 0.106 & 0.244 \\
Avg.\ leaf depth     & 0.342 & 0.000 & 0.084 & 0.521 \\
Branching factor     & 0.111 & 0.200 & 0.062 & 0.858 \\
\midrule
Features passing & \multicolumn{2}{c}{$2/8$} & \multicolumn{2}{c}{$8/8$} \\
\bottomrule
\end{tabular}
\end{table}

Table~\ref{tab:ops} compares operator usage between the corpus and the two generators.
The B-PCFG tracks the corpus distribution more closely across the common operators and reproduces rarer ones such as the trigonometric and exponential functions, whereas both generators fail to emit the least frequent operators (arcsin and tanh).

\begin{table}[H]
\centering
\caption{Operator frequencies (\%) in the Feynman corpus and in $1000$ equations generated by the standard PCFG and the optimised B-PCFG.}
\label{tab:ops}
\begin{tabular}{lccc}
\toprule
Operator & Corpus & PCFG & B-PCFG \\
\midrule
Pow    & 47.1 & 41.3 & 39.6 \\
Mul    & 33.2 & 41.9 & 31.4 \\
Add    & 12.6 & 10.5 & 17.9 \\
cos    &  2.3 &  2.3 &  4.8 \\
exp    &  2.1 &  1.5 &  2.7 \\
sin    &  1.9 &  2.3 &  3.3 \\
arcsin &  0.4 &  0.0 &  0.0 \\
log    &  0.2 &  0.1 &  0.4 \\
tanh   &  0.2 &  0.0 &  0.0 \\
\bottomrule
\end{tabular}
\end{table}

\subsection{Hyperparameter Ranking Task}
\begin{figure}[H]
  \centering
  \includegraphics[width=1\linewidth]{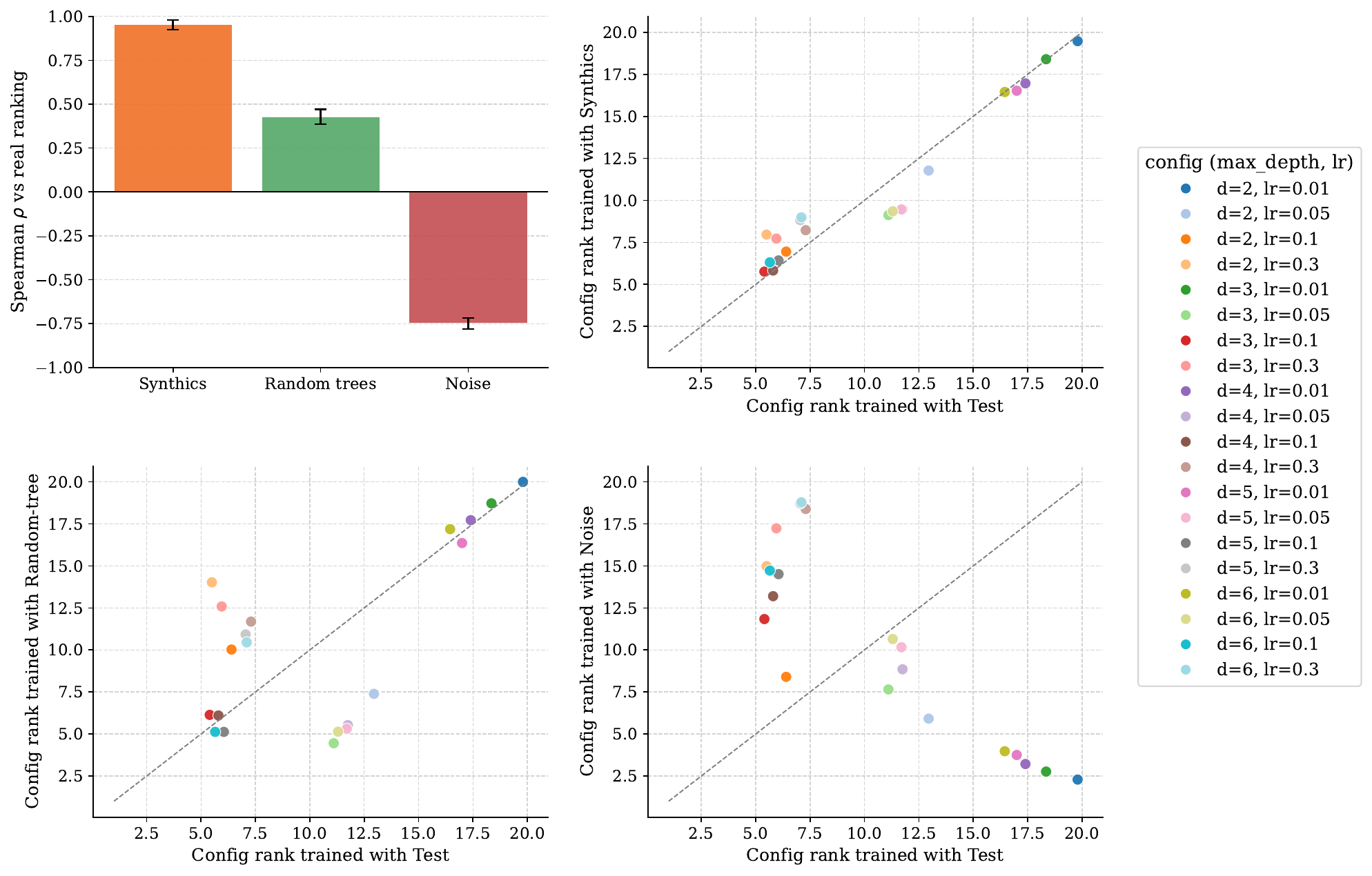}
  \caption{Average configuration rankings across tasks for each source.
  The real-data ranking is the baseline, and the correlation of each other source's average ranking with the real-data ranking is reported as $\rho_S$.
  In addition, the configuration ranking for each source is compared with the real-data ranking.
  The closer the agreement of configurations, the more the points lie on the diagonal.}
  \label{fig:config_rankings}
\end{figure}

The hyperparameter ranking task assesses the practical utility of the synthetic datasets for guiding model selection in real machine-learning tasks.
Figure~\ref{fig:config_rankings} shows the average configuration rankings across tasks for each source.
It can be seen that the Synthics (B-PCFG) source produces a configuration ranking that is closely aligned with the real-data ranking, as reflected in the high Spearman correlation.
Random trees also produce a ranking that is somewhat correlated with the real-data ranking.
The Noise source produces a ranking that is essentially uncorrelated with the real-data ranking, as expected.

\begin{figure}[H]
  \centering
  \includegraphics[width=0.65\linewidth]{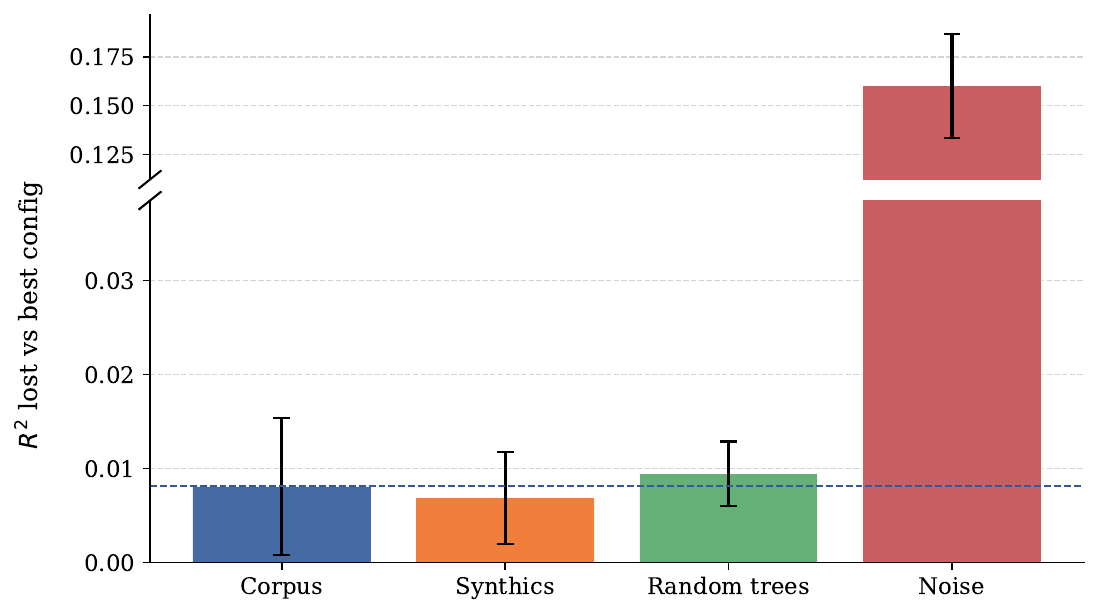}
  \caption{The regret of each source, measured as the average $R^2$ loss compared to the real-optimal configuration when applying the synthetic-optimal configuration to the real tasks.}
  \label{fig:regret}
\end{figure}

Based on the best configuration determined by each source's ranking, the average $R^2$ loss compared to the real-optimal configuration is computed across all tasks, which serves as the regret of that source.
Table~\ref{tab:deployment_real_data} shows the regret and pick rank of each source.
The pick rank indicates the relative ordering of the best configuration determined by each source, as an average across the 5 random seeds and 20 test tasks.
Synthics is able to pick a configuration that is on average the 6th best out of 20, which is a substantial improvement over random trees and noise, which pick configurations that are on average the 10th and 19th best respectively.
For the regret, Synthics achieves similar accuracy to that of the real data. However, so does random trees.
This suggests that the top of the real-data ranking is relatively flat, with many configurations performing similarly, so that even picking a configuration that is on average the 10th best can still yield low regret.
For a visual comparison of the regret across sources, see Figure \ref{fig:regret}.

\begin{table}[H]
\centering
\caption{Deployment metrics on independent real data across four different data sources.
Regret measures the average $R^2$ performance loss compared to the real-optimal configuration, while pick rank indicates the relative ordering out of 20 total configurations.
Note that the ordering is averaged across 5 random seeds, so the pick rank is therefore unlikely to be as low as 1, even for the real data.}
\label{tab:deployment_real_data}
\begin{tabular}{lcc}
\toprule
 & \multicolumn{2}{c}{Deployment Metrics} \\
\cmidrule(lr){2-3}
Source & Regret ($R^2$ loss) & Pick rank \\
\midrule
Corpus        & 0.008 & \phantom{0}6.4 / 20 \\
Synthics      & 0.007 & \phantom{0}6.0 / 20 \\
Random trees  & 0.009 & \phantom{0}9.9 / 20 \\
Noise         & 0.160 & 19.2 / 20 \\
\bottomrule
\end{tabular}
\end{table}

\section{Discussion}

The structural comparison isolates the role of the Bayesian smoothing, since the two generators (PCFG and B-PCFG) differ only in the concentration parameter $\alpha$.
With the soft-forcing temperature held at its optimised value, adding the prior raised the number of features consistent with the corpus from two to eight.
The temperature itself remains important, as it governs the depth of the generated trees, but at a well-chosen temperature it is the smoothing that closes the remaining structural gap: without it the raw maximum-likelihood grammar places most of its probability on the few most frequent production rules and collapses toward shallow, simple expressions, whereas the smoothed grammar recovers the longer tail of deeper and more varied trees seen in the corpus.

The strength of prior required is likely to depend on the size of the corpus.
With only $100$ equations the empirical rule frequencies are sparse and noisy, so a strong Dirichlet prior is needed to fill in plausible but unobserved expansions.
For a substantially larger corpus the maximum-likelihood frequencies should already be well estimated, and heavy smoothing may become unnecessary or even detrimental.
This conjecture is left to future work.

There are a few important limitations in the presented method and results.
Passing a KS test indicates only that the corpus and generated distributions cannot be distinguished at the present sample size, not that they are identical.
Several features clear the threshold only marginally and the histograms remain visibly different in places, so a larger generated sample could expose residual mismatch.
The method is also restricted to closed-form algebraic equations and does not address ordinary or partial differential equations, and the reported outcome is both seed-dependent and validated on a single corpus.

It was shown in the hyperparameter ranking task that the synthetic datasets generated by the optimised B-PCFG can guide model hyperparameter selection for real machine-learning tasks, outperforming random trees and noise.
The regret of the Synthics source is, however, similar to that of random trees, a consequence of the flat top of the real-data ranking discussed in the Results.
Also, the hyperparameter ranking task is only a simple example of a practical application, and more comprehensive evaluation across different tasks and models would be needed to fully establish the utility of the synthetic datasets.

Despite these limitations, structurally faithful synthetic equations offer a practical route to training data where real measurements are scarce, as in many engineering domains, and could serve as priors for models that learn directly from synthetic distributions such as TabPFN~\citep{hollmann2025accurate}.
Natural extensions include generating differential equations, validating the approach on larger and more diverse corpora, and, most importantly, more comprehensive evaluation of whether models trained on the synthetic data transfer to real machine-learning tasks.

\section{Conclusions}

This paper introduced Synthics, a method that learns a Bayesian Probabilistic Context-Free Grammar from a corpus of physics equations and generates novel regression datasets that preserve the structural properties of the corpus, combining domain-aware input sampling with Kolmogorov--Smirnov validation.
Bayesian smoothing proved essential: the optimised B-PCFG produced equation distributions statistically consistent with the corpus on all eight structural features, compared with two for an unsmoothed grammar.
It was shown that the B-PCFG-generated datasets can guide hyperparameter tuning in a simple machine-learning task, providing a practical validation of the synthetic data.
The present work is limited to closed-form algebraic equations.
Ordinary and partial differential equations, which describe a large part of engineering and the physical sciences, fall outside the current grammar and are left to future work.
Future research should also focus on validation on larger corpora and testing on more advanced learning tasks.

\section*{Acknowledgements}
During the preparation of this work, the author used Claude Code (Sonnet 4.6 and Opus 4.7) to assist with code development, text writing, and editing.
The AI search engine Consensus was also used for literature discovery.
The underlying research, methodology, and conclusions are entirely original and were conceived and executed by the author.

\section*{Code}
The code for the Synthics data generator is available at \url{https://github.com/vepsalai/synthics}.

\bibliographystyle{unsrtnat}
\bibliography{references}

\appendix
% --- Add these lines right here ---
\setcounter{figure}{0}
\renewcommand{\thefigure}{A\arabic{figure}}
% ----------------------------------

\newpage
\section*{Appendix}

\begin{figure}[H]
\centering
\includegraphics[width=\linewidth]{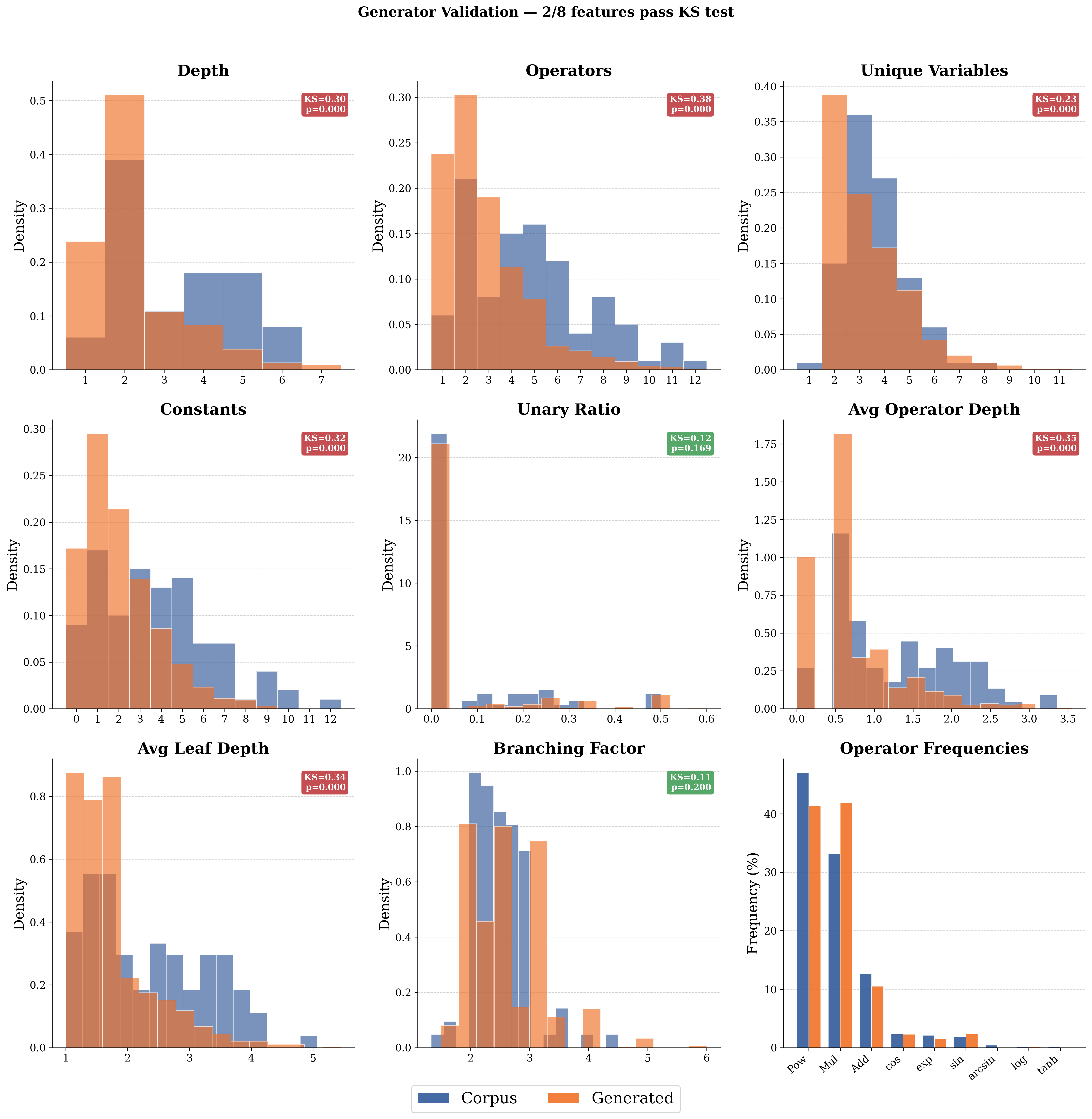}
\caption{Validation of the standard PCFG. Per-feature distribution comparison between the Feynman corpus (blue) and $1000$ generated equations (orange); the final panel shows operator frequencies. Two of the eight structural features pass the KS test.}
\label{fig:pcfg_validation}
\end{figure}

\begin{figure}[H]
\centering
\includegraphics[width=\linewidth]{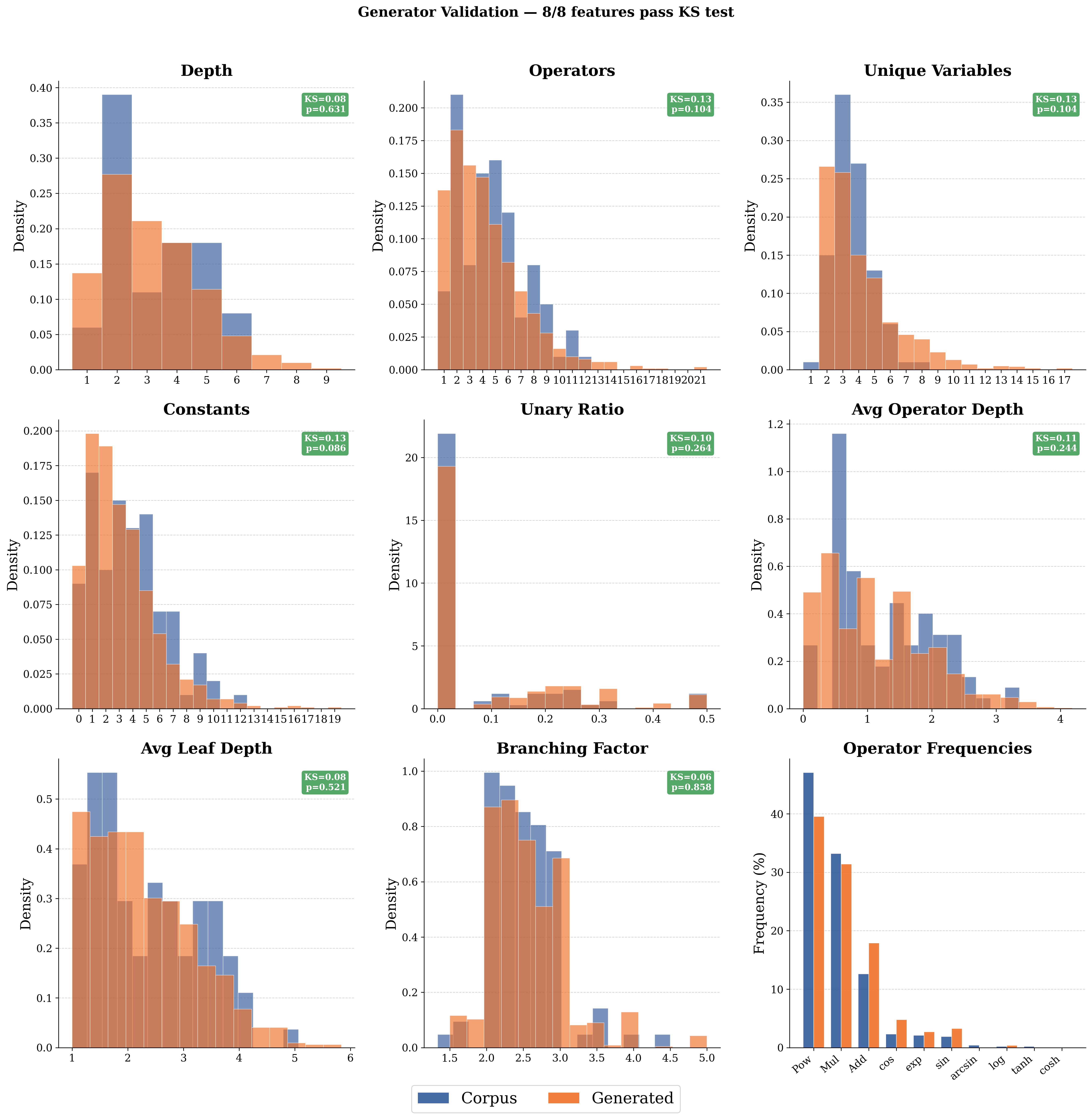}
\caption{Validation of the optimised B-PCFG ($\alpha^* = 44$, $\tau^* = 6$). Per-feature distribution comparison between the Feynman corpus (blue) and $1000$ generated equations (orange); the final panel shows operator frequencies. All eight structural features pass the KS test.}
\label{fig:bpcfg_validation}
\end{figure}

\end{document}